\documentclass[10pt,twocolumn,letterpaper]{article}

\usepackage{wacv}
\usepackage{times}
\usepackage{epsfig}
\usepackage{graphicx}
\usepackage{amsmath}
\usepackage{amssymb}
\usepackage{amsmath}

\usepackage[pagebackref=true,breaklinks=true,letterpaper=true,colorlinks,bookmarks=false]{hyperref}

\wacvfinalcopy 

\def\wacvPaperID{342} 

\ifwacvfinal\fi
\begin{document}

\title{MAC: Mining Activity Concepts for Language-based Temporal Localization}

\author{Runzhou Ge \quad Jiyang Gao \quad Kan Chen \quad Ram Nevatia \\
University of Southern California \\
{\tt\small \{rge, jiyangga, kanchen, nevatia\}@usc.edu} 
}


\maketitle
\ifwacvfinal\thispagestyle{empty}\fi

\begin{abstract}

We address the problem of language-based temporal localization in untrimmed videos.
Compared to temporal localization with fixed categories, this problem is more challenging as the language-based queries not only have no pre-defined activity list but also may contain complex descriptions.
Previous methods address the problem by considering features from video sliding windows and language queries and learning a subspace to encode their correlation, which ignore rich semantic cues about activities in videos and queries. 
We propose to mine activity concepts from both video and language modalities by applying the actionness score enhanced Activity Concepts based Localizer (ACL). 
Specifically, the novel ACL encodes the semantic concepts from verb-obj pairs in language queries and leverages activity classifiers' prediction scores to encode visual concepts. 
Besides, ACL also has the capability to regress sliding windows as localization results.
Experiments show that ACL significantly outperforms state-of-the-arts under the widely used metric, with more than 5\% increase on both Charades-STA and TACoS datasets\footnotemark.

\end{abstract}

\section{Introduction}


\footnotetext{Code is available in  
\url{https://github.com/runzhouge/MAC}}

We address natural language-based temporal localization in untrimmed videos.
Given an untrimmed video and a natural language query, natural language-based temporal localization aims to determine the start and end times of the activities in the query. Compared to the activity temporal localization task~\cite{Shou_2017_CVPR, gao2017cascaded, Xu_2017_ICCV, Zhao_2017_ICCV, Chao_2018_CVPR}, the natural language-based temporal localization has two unique characteristics. First, the natural language-based temporal localization is not constrained by a pre-defined activity label list; second, the language query can be more complex and may contain multiple activities. \eg ``person begin opening the refrigerator to find more food''.

Recent work~\cite{shou2016temporal, Gao_2017_ICCV, Hendricks_2017_ICCV, liu2018attentive, wu2018multi} tries to solve this problem through learning correlations between the query and the video. Specifically, they use a pre-trained Convolutional Neural Network (CNN) ~\cite{Tran_2015_ICCV, he2016deep, Tran_2018_CVPR} and Long Short Term Memory (LSTM) network~\cite{hochreiter1997long} to get the video clip features and query embeddings respectively; then, the video features and query embeddings are mapped to a common space. In this common space, Hendricks~\etal~\cite{Hendricks_2017_ICCV} measure the squared distance between two modalities. Others~\cite{Gao_2017_ICCV, liu2018attentive, wu2018multi} use a Multi-Layer Perceptron (MLP) to get the matching score after exploring the possible interactions between language and video.

\begin{figure}
\begin{center}
\includegraphics[width=0.47\textwidth]{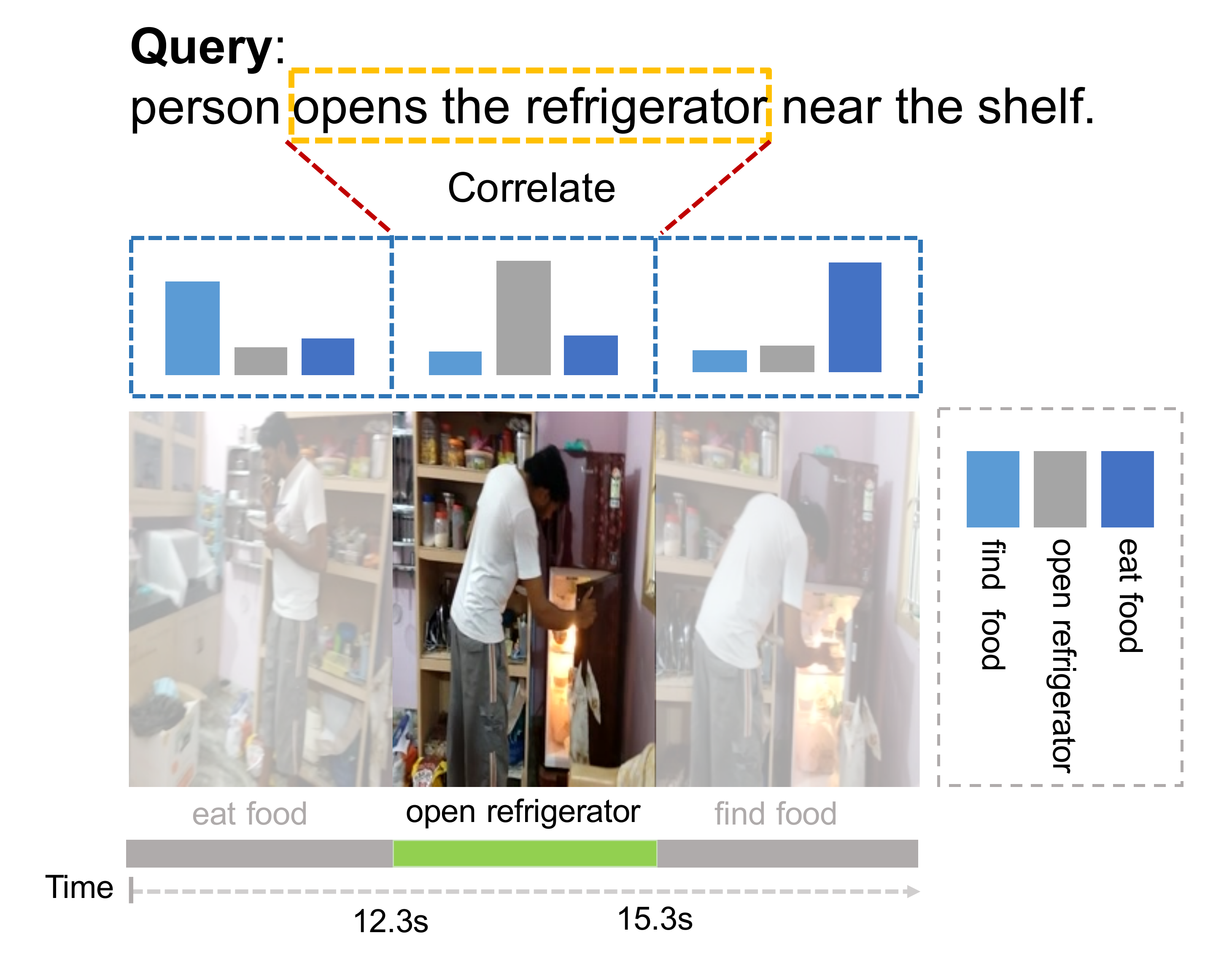}
\end{center}
\caption{We mine the activity concepts from both the language query and the video. \emph{verb-obj} pairs are used as the sentence activity concepts (in yellow dashed box). The probability distribution of pre-defined activity labels is used as the visual activity concepts (the histogram in blue dashed boxes).}
\label{fig:fig1}
\end{figure}


Previous methods~\cite{Gao_2017_ICCV, Hendricks_2017_ICCV, liu2018attentive, wu2018multi} calculate the visual-semantic correlation based on the whole sentence and the video clip, activity information is not modelled explicitly on both the language and video modalities. On the language side, they are \emph{verb-obj} pairs (\eg ride bike) embedded in the queries (an example is shown in Figure \ref{fig:fig1}). These \emph{verb-obj} pairs (VO) contain rich activity information, and can be viewed as \emph{semantic activity concepts}; on the video side, when we extract video features, the last layer outputs of the pre-trained CNN \cite{Tran_2015_ICCV, he2016deep, Tran_2018_CVPR} are usually activity labels. These class probability distribution represents the \emph{visual activity concepts}. \emph{Mining} such activity concepts from video and query and using them in visual-semantic correlation can be helpful. 

\cite{Gao_2017_ICCV, liu2018attentive, wu2018multi} use sliding windows to scan the video without any selection bias during test. But actually there are many more background video clips in these sliding windows which contain nothing meaningful inside.
It's more desirable to have an actionness confidence score for each sliding window, which could reflect how confident the sliding window contains activities. 
We propose a novel actionness score enhanced Activity Concepts based Localizer (ACL) to mine the activity concepts from both the video and language modality and to utilize the actionness score for each sliding window. 
To mine the activity concepts, VO are parsed from the sentence query. \eg ``open refrigerator'' in Figure \ref{fig:fig1}; its embedding is used as the semantic activity concept. 
Visual activity concept comes from probability distribution of pre-defined activity labels from the classifier level. Note that the query activity need not be in the set of pre-defined labels, we use the distribution as yet another feature vector and experiments described later show its utility.
We separately process the pair of activity concepts, and the pair of visual features and sentence embeddings.
The concatenation of two multi-modal processing outputs are fed into a two-layer MLP which outputs alignment score and regressed temporal boundary. We find that including actionness scores in the final stage helps improve the alignment accuracy. 


Our contributions can be summarized as below:

(1) We propose to mine the activity concepts from both videos and language queries to tackle language-based temporal localization task.

(2) We introduce an actionness score enhanced pipeline to enhance the localization.

(3) We validate the effectiveness of our proposed model on two datasets and achieve the state-of-the-art  performance on both.

In the following, we first discuss related work in Section \ref{related_work}. More details of ACL are provided in Section \ref{methods}. Finally, we analyze and compare ACL with other approaches in Section \ref{evaluation}.

\section{Related Work}
\label{related_work}
\textbf{Localization by Language} aims to localize mentioned activities / objects in video / image modalities by a natural language query. 

For video modality, Gao \etal~\cite{Gao_2017_ICCV} propose a cross-modal temporal regression localizer (CTRL) to jointly model language query and video clips in long untrimmed videos. The CTRL not only outputs matching score and but also the boundary offsets to refine the input video clip's location. Hendricks \etal ~\cite{Hendricks_2017_ICCV} apply moment context network to transform the visual features and sentence embeddings to a common space and measured the distance between two modalitites to find the best matching video segments for each sentence query. Liu \etal~\cite{liu2018attentive} apply a memory attention mechanism to emphasize the visual features mentioned by the query and simultaneously incorporate the their context. Wu and Han~\cite{wu2018multi} improve the localizing performance by plugging in the Multi-modal Circulant Fusion module to CTRL~\cite{Gao_2017_ICCV}. To be specific, they try to explore more possible interactions between vectors of different modalities in cross-modal processing. 

For image modality, Hu \etal~\cite{hu2016natural} propose to localize mentioned objects by reconstructing language queries in weakly supervised scenario. Chen \etal~\cite{chen2018msrc,chen2017query} adopt a regression mechanism in supervised scenario. Recently, Chen \etal~\cite{chen2018knowledge} introduce visual consistency to further boost the performance in weakly-supervised localization. 

\textbf{Concept Discovery} is a technique to discover the concepts from image-sentence corpora~\cite{sun2013active, gao2016acd}. 
Sun \etal~\cite{sun2015automatic} design a visual concepts discovering and clustering algorithm for general objects and validate it on the task of bidirectional image and sentence retrieval. 
Gao \etal~\cite{gao2016acd} extend the general objects visual concepts to action concepts and achieve good performance on weakly supervised action classification task in still images.
For temporal activity localization, we further leverage concepts from both video and language modalities to find mentioned activities more accurately.

\textbf{Temporal Activity Proposal Generation} aims to generate high quality proposals containing actions with high recall, which can be used for activity detection and localization in later stages~\cite{gao2017cascaded,shou2016temporal,bolles20142014}. To address the problem, Gao \etal~\cite{TURN_2017_ICCV} cut videos into units and adopt a regression mechanism (TURN) to generate and refine temporal activity proposals' boundaries. Zhao \etal~\cite{zhao2017temporal} leverage snippet level actionness score and apply a temporal action grouping (TAG) algorithm to generate activity proposals. Recently, Gao \etal~\cite{gao2018ctap} introduce complementary filtering to combine advantages of TURN~\cite{TURN_2017_ICCV} and TAG~\cite{zhao2017temporal} to further boost the quality of generated temporal activity proposals. Inspired by previous work, we apply TURN~\cite{TURN_2017_ICCV} as the temporal proposal generator for later temporal activity localization.

\begin{figure*}
\begin{center}

\includegraphics[width=6.8in]{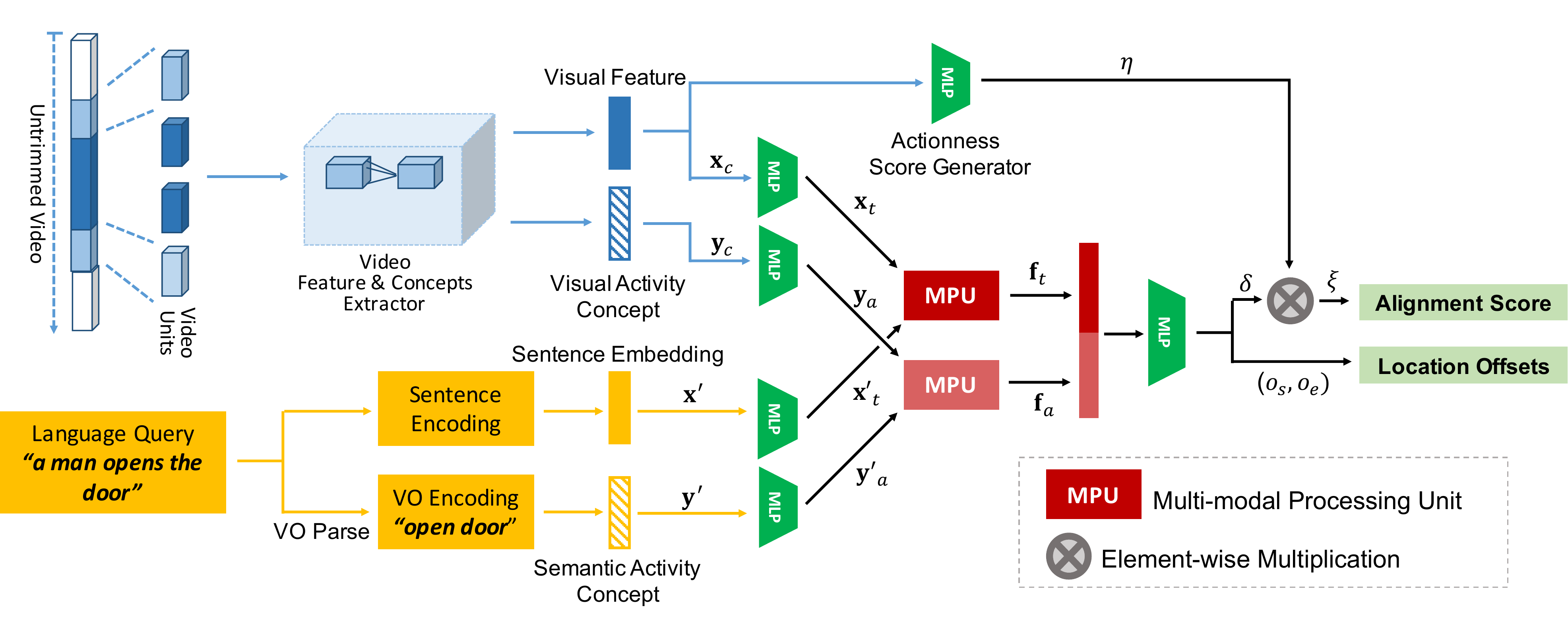}
\end{center}
\caption{The framework of actionness score enhanced Activity Concepts based Localizer (ACL) pipeline. The whole pipeline consists of video pre-processing, correlation-based alignment, activity concepts mining and actionness score enhanced localization.}
\label{fig:general}
\end{figure*}



\section{Methods}
\label{methods}
In this section, we present the details of actioness score enhanced Activity Concepts based Localizer (ACL) from four aspects: video pre-processing, correlation-based alignment, activity concepts mining and actionness score enhanced localization. Then we provide details of model training. The framework is shown in Figure~\ref{fig:general}.


\subsection{Video Pre-processing}
We follow the method of~\cite{TURN_2017_ICCV} to process the videos to get the unit-level visual features.
Video units are used to avoid repeated feature extraction. 
A video is first decomposed into consecutive video units with frame length $l_u$. The video unit $u$ is used to extract unit-level feature through a feature extractor $\mathcal{E}_v$ which can be expressed as $\mathbf{x}_i=\mathcal{E}_v(u)$. These unit-level features serve as the basic feature unit for clip-level features. 
\label{subsec:mult_laern}

\subsection{Correlation-based Alignment}
\label{subsec:corr_learning}
\textbf{Clip-level Visual Feature}. 
Video clips for a given video are generated by sliding windows. For a video clip $c$, we consider itself as the central clip and its surrounding clips as pre- and post-contextual clips. Following~\cite{TURN_2017_ICCV}, a central clip $\{u_i\}_{s_u}^{s_u+n_u}$ contains $n_u$ consecutive units, where $s_u$ is the index of the start unit, and $e_u=s_u+n_u$ is the index of the end unit. The pre-contextual clip $\{u_i\}_{s_u-1-n_{ctx}}^{s_u-1}$ and post-contextual clip $\{u_i\}_{e_u+1}^{e_u+1+n_{ctx}}$ are the contextual clip right before and after the central clip, where $n_{ctx}$ is a hyper parameter denoting the number of units that the contextual clip has. To get the clip-level features for both the central clips and the contextual clips, we apply the pooling operation, $\mathcal{P}$, to every unit feature. The feature for video clip $c$ is
\begin{equation}
\small
\mathbf{x}_c = \mathcal{P}({\{\mathbf{x}_i\}_{s_u-1-n_{ctx}}^{s_u-1}}) 
\parallel \mathcal{P}({\{\mathbf{x}_i\}_{s_u}^{e_u}}) 
\parallel \mathcal{P}({\{\mathbf{x}_i\}_{e_u+1}^{e_u+1+n_{ctx}}})
\label{eqn:vis_feats}
\end{equation}
where $\parallel$ denotes vector concatenation. We use average pooing as $\mathcal{P}$. After that, we use a fully-connected ($\mathit{fc}$) layer to project it to vector $\mathbf{x}_t$ with dimension $d_t$.

\textbf{Query Embedding}. Similar to the visual feature, we use a query encoder $\mathcal{E}_s$ to get the embedding of a query $q$, which can be exprssed as $\mathbf{x}'=\mathcal{E}_s(q)$. Then we linearly map the query embedding from $\mathbf{x'}$ to $\mathbf{x}'_t$ with dimension $d_t$. 

\textbf{Multi-modal Processing Unit (MPU)}. We adopt the multi-modal processing method~\cite{Gao_2017_ICCV} to compute the interactions between two vectors of the same length. Here it takes vector $\mathbf{x}_t$ and $\mathbf{x}'_t$ as inputs. Element-wise multiplication ($\otimes$) and element-wise addition ($\oplus$) and vector concatenation ($\parallel$) are used to explore interactions between different modalities. Then we concatenate three operation results
\begin{equation}
\mathbf{f}_t = (\mathbf{x}_t \otimes \mathbf{x}'_t)
\parallel (\mathbf{x}_t \oplus \mathbf{x}'_t)
\parallel (\mathbf{x}_t \parallel \mathbf{x}'_t)
\end{equation}
where $\mathbf{f}_t$ is the outputs of multi-modal processing. Since the element-wise operation doesn't change the vector dimension, the dimension of $\mathbf{f}_t$ is $4d_t$. Multi-modal processing is not only used in learning correlations between visual features and sentence embeddings but also used in learning correlations between visual activity concepts and semantic activity concepts. We denote the outputs of activity concept multi-modal processing $\mathbf{f}_a$ in advance.

\textbf{Alignment}.
The outputs of multi-modal processing $\mathbf{f}_t$ are fed into the MLP network to get an alignment score for clip $c$ and query $q$ (as also in~\cite{Gao_2017_ICCV, wu2018multi}). 


\subsection{Activity Concepts}

We leverage activity concepts from both videos and queries to facilitate activity localization.

\textbf{Semantic Activity Concept}. We use the \emph{verb-obj} pairs (VO) as the semantic activity concept. We first parse the sentence and get a two-word VO of each sentence. Then we lemmatize every word in VO to get the normal form of VO. \eg ``person opens the refrigerator near the shelf.'' is first parsed to get ``opens refrigerator'' and then do the lemmatization to get ``open refrigerator'' as VO. A word encoder $\mathcal{E}_d$ is used to get the word embedding for each word. Two words' embeddings are concatenated as the VO embedding $\mathbf{y}'$ and linearly projected to $d_a$-dimension $\mathbf{y}'_a$ through a $\mathit{fc}$ layer for the activity multi-modal processing. 

\textbf{Visual Activity Concept}. For feature, we use the outputs of the $\mathit{fc}6$ layer as the features in C3D~\cite{Tran_2015_ICCV}.
These features contain holistic information of the video but are without any semantic meaning. The outputs of the $\mathit{fc}8$ layer 
could be expressed as the unnormalized distribution of pre-defned activity labels. In activity classification~\cite{Tran_2015_ICCV}, these outputs are used as the prediction of the video activity.
We do not use them to classify activities directly but, instead, use the distribution as a semantic feature that is helpful in classifying the query activity.


Specifically, the outputs $\mathbf{y}_i$ of $\mathit{fc}8$ can be treated as unit-level visual activity concepts which have the same size as the pre-defined number of activity labels. Following Section \ref{subsec:mult_laern}, the clip-level visual activity concept are pooled from unit-level concept as
\begin{equation}
\mathbf{y}_c = \mathcal{P}({\{\mathbf{y}_i\}_{s_u}^{e_u}}) 
\end{equation}
To avoid the activity noise from the contextual clips, we only use the central clip to get the clip-level visual activity concept. The $\mathbf{y}_c$ is linearly projected to $\mathbf{y}_a$ for activity concept multi-modal processing. $\mathbf{y}_a$ is of dimension $d_a$.

\textbf{Activity Concept Mutli-modal Processing}. The semantic and visual activity concepts are mined from videos and sentence queries separately. Although they both contain the activity concepts, they are in different domains. To explore the interactions between two activity concepts, we conduct the multi-modal procssing to the two linearly projected concepts, which is similar to Section~\ref{subsec:corr_learning}. 
\begin{equation}
\mathbf{f}_a = (\mathbf{y}_a \otimes \mathbf{y}'_a)
\parallel (\mathbf{y}_a \oplus \mathbf{y}'_a)
\parallel (\mathbf{y}_a \parallel \mathbf{y}'_a)
\end{equation}
So the outputs $\mathbf{f}_a$ is with dimension $4d_a$.



\subsection{Actionness Score Enhanced Localization}
To give each sliding window a score of how confident the sliding window contains activities in the process of generating test samples, we design an actionness score enhanced localization pipeline. During test, it generates the actionness score for every sliding window candidate. 

\textbf{Actionness Scores}. The actionness score generator is a two-layer MLP. The generator is trained separately to have the capability to give sliding window candidate $c_i$ a confidence score $\eta_i$ . The confidence score $\eta_i$ indicates ``the likelihood of the sliding window candidate containing meaningful activities''.

\textbf{Temporal Localization}. Previous localization methods~\cite{Gao_2017_ICCV, wu2018multi} only take the correlation of complete sentence embedding and visual feature $\mathbf{f}_t$ as inputs. Our localization network is similar to~\cite{Gao_2017_ICCV} but takes the concatenation of outputs of multi-modal processing outputs, $\mathbf{f}_t \parallel \mathbf{f}_a$, as input. It generates pre-alignment score $\delta_{i,j}$ and regression offsets $(o_s, o_e)$, where $o_s$ is the start video unit offset, $o_e$ is the end video unit offset. We multiply pre-alignment score $\delta_{i,j}$ with the actionness score $\eta_i$ to get the final alignment score $\xi_{i,j}$. The alignment score $\xi_{i,j}$ is used as the final score to predict the alignment confidence between clip $c_i$ and query $q_j$. The regression offsets $(o_s, o_e)$ are used to refine the clip temporal location.


\subsection{Model Training}
The actionness score generator and ACL are trained separately. 


\textbf{Actionness Score Generator Training}. To train the score generator, we assign a binary class label (non-background/background) to each clip candidate following~\cite{TURN_2017_ICCV}. We use a binomial cross-entropy loss to train the generator.

\textbf{ACL Training}. The temporal localization network has two outputs, the pre-alignment score $\delta_{i,j}$ and the regression offsets $(o_s, o_e)$. A multi-task loss $\mathcal{L}_{loc}$ is used to train for the pre-alignment and the localization regression. 
\begin{equation}
\mathcal{L}_{loc} = \mathcal{L}_{aln}+\beta \mathcal{L}_{rgr}
\end{equation}
where $\mathcal{L}_{aln}$ is the pre-alignment loss and $\mathcal{L}_{rgr}$ is the location regression loss. $\beta$ is a hyper parameter. Specifically, the $\mathcal{L}_{aln}$ is
\begin{equation}
\mathcal{L}_{aln} = \frac{1}{N}\sum_{i=1}^{N}\Big[\gamma \mathrm{log}(1+e^{-\delta_{i,j}})+\sum_{j=0,j\neq i}^{N} \mathrm{log}(1+e^{\delta_{i,j}})\Big]
\end{equation}
where $\gamma$ is the a hyper parameter to control the weights between positive and negative samples, 
$\delta_{i,j}$ is hte pre-laignment score between clip $c_i$ and query $q_j$, $N$ the batch size.
\begin{equation}
\mathcal{L}_{rgr} = \frac{1}{N}\sum_{i=1}^{N}\Big[\mathcal{S}(o^*_{s,i}-o_{s,i})+\mathcal{S}(o^*_{e,i}-o_{e,i})\Big]
\end{equation}
where $\mathcal{S}$ is the smooth $L_1$ loss function, $*$ denotes the ground truth.

The rectified linear unit (ReLU) is selected as the non-linear activation function and Adam~\cite{kingma2014adam} algorithm is used for optimization.




\section{Evaluation}
\label{evaluation}

We evaluate our model on Charades-STA~\cite{Gao_2017_ICCV} and TACoS~\cite{regneri2013grounding}. In this section, we describe evaluation settings and discuss the experiment results.

\subsection{Datasets}

\begin{table}
\begin{center}
\begin{tabular}{l|cc}
\hline
Datasets            &Charades-STA~\cite{Gao_2017_ICCV}      &TACoS~\cite{regneri2013grounding}   \\
\hline\hline
\# Videos               &6,672                              &127    \\
\# Clips                &11,772                             &3,290    \\
\# Queries         &16,128                             &18,818     \\ 
Avg. \# Clips            &1.76                               &25.91  \\
Avg. \# Queries           &1.37                               &5.72   \\
Avg. \# Words           &6.22                               &8.87    \\
$\mathit{\sigma}^2$ &3.65                             &33.87   \\

\hline
\end{tabular}
\end{center}
\caption{Statistics of Charades-STA and TACoS datasets. ``\#'' stands for number. $\mathit{\sigma}^2$ is the variance of number of words in each query.} 
\label{tbl:dataset}
\end{table}

\begin{table}[t]
\begin{center}
\setlength\tabcolsep{4pt}
\begin{tabular}{l|cccc}
\hline
Method  
&\begin{tabular}{@{}c@{}}$\mathrm{R@1}$ \\ $\mathit{IoU}$=0.7\end{tabular}
&\begin{tabular}{@{}c@{}}$\mathrm{R@1}$ \\ $\mathit{IoU}$=0.5\end{tabular}
&\begin{tabular}{@{}c@{}}$\mathrm{R@5}$ \\ $\mathit{IoU}$=0.7\end{tabular}
&\begin{tabular}{@{}c@{}}$\mathrm{R@5}$ \\ $\mathit{IoU}$=0.5\end{tabular}\\

\hline\hline
CTRL\footnotemark~\cite{Gao_2017_ICCV}
&7.15  &21.42  &26.91  &59.11\\
\hline
Swin+Score
&8.98     &22.20     &29.38  &58.06\\
Prop+Score
&\textbf{9.14}              &\textbf{22.63}              &\textbf{30.08}  &\textbf{59.23}\\

\hline
\end{tabular}
\end{center}
\caption{Evaluation of actionness score on Charades-STA.}
\label{tbl:score}
\end{table}

\textbf{Charades-STA}~\cite{Gao_2017_ICCV} was built on Charades~\cite{sigurdsson2016hollywood}. Charades contains 9,848 videos and was first collected from everyday activities for activity understanding. 
Gao~\etal~\cite{Gao_2017_ICCV} enhanced it to make it suitable for language-based temporal localization task.

\textbf{TACoS} is from MPII Composites dataset~\cite{rohrbach2012script} which contains different activities in the cooking domain. Regneri~\etal~\cite{regneri2013grounding} extended the natural language descriptions by crowd-sourcing. They also aligned each descriptions to its corresponding video clips.  
More details of Charades-STA and TACoS are shown in Table~\ref{tbl:dataset}.

\footnotetext{The author did some cleaning to the dataset. Updated results could be found in \url{https://github.com/jiyanggao/TALL}}


\begin{figure}[t]
\begin{center}
\includegraphics[width=0.47\textwidth]{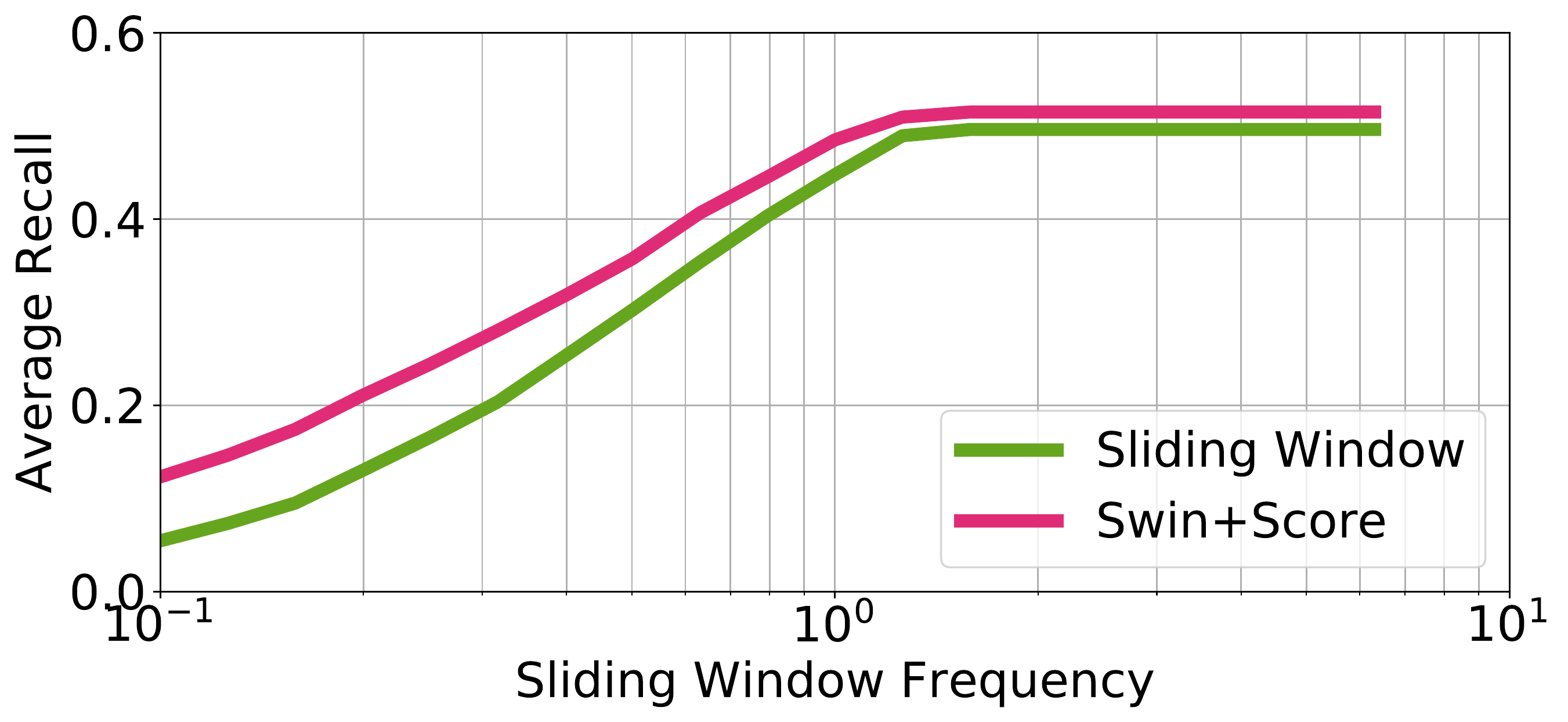}
\end{center}
\caption{Comparison of our Swin+Score method and purely sliding window method under the AR-F metric on Charades-STA.}
\label{fig:ARF}
\end{figure}

\begin{figure}[t]
\begin{center}
\includegraphics[width=0.47\textwidth]{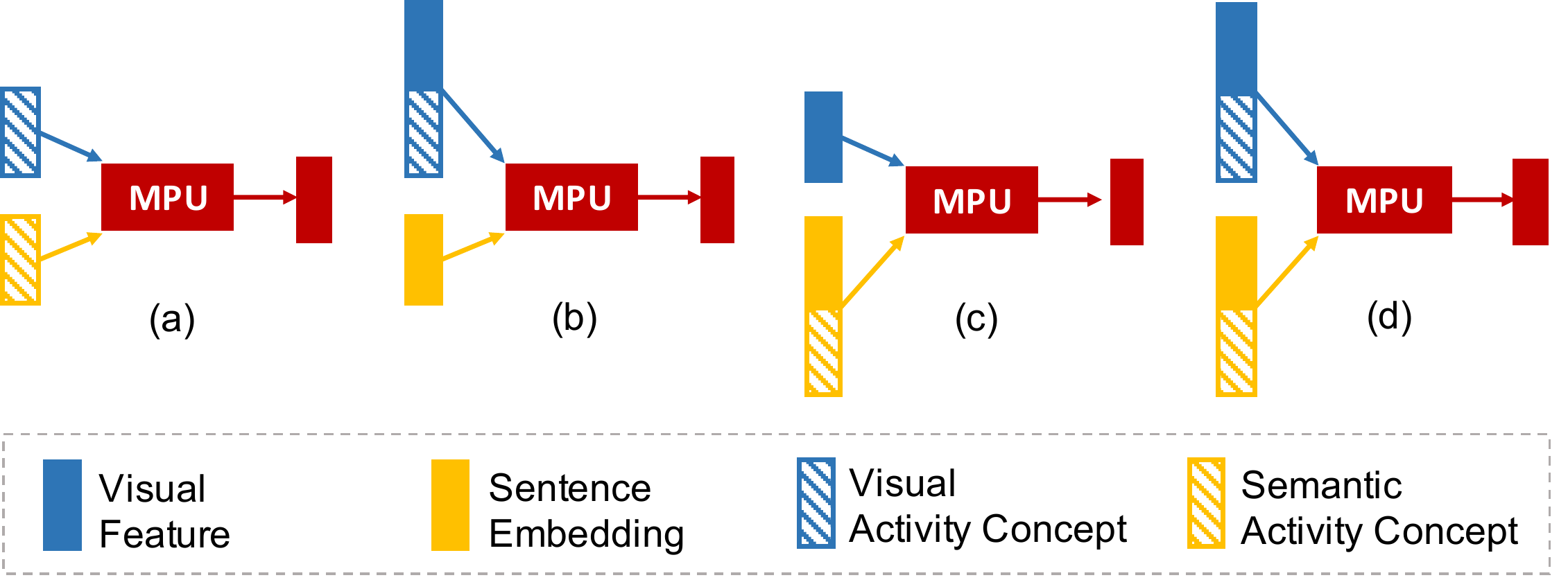}
\end{center}
\caption{System variants. (a) Activity only; (b) w/o SAC; (c) w/o VAC; (d) Concat.}
\label{fig:variant}
\end{figure}

\subsection{Experiment Setup}
\label{subsec:exp_setup}
\textbf{Visual Features}. 
The videos are first decomposed into units with 16 consecutive frames which serve as basic processing blocks. We use the C3D~\cite{Tran_2015_ICCV} pre-trianed on Sports-1M~\cite{KarpathyCVPR14} dataset to process the frames in the unit. The outputs of C3D $\mathit{fc6}$ layer ($\in{\mathbb {R}^{4,096}}$) are used as the unit-level visual features.
The contextual clip length is set to 128 frames and the number of contextual clip is set to 1 in our work. (we average 8 consecutive unit-level features before and after the central clip to get the pre-contextual and post-contextual clip features, respectively). 
According to Equation \ref{eqn:vis_feats}, the dimension of visual feature is 12,288.

\textbf{Visual Activity Concept}. We use the outputs of $\mathit{fc}8$ layer ($\in{\mathbb {R}^{487}}$)  of C3D~\cite{Tran_2015_ICCV} as the visual activity concept from the video. The C3D~\cite{Tran_2015_ICCV} was pre-trained on Sports-1M~\cite{KarpathyCVPR14} which has 487 sports activity classes. 
Since Sports-1M only contains the sports actions, we also explore Kinetics~\cite{kay2017kinetics} dataset which has more common human actions and which has a larger activity label overlap with the Charades-STA~\cite{Gao_2017_ICCV} and TACoS~\cite{regneri2013grounding}. We use the outputs of the last $\mathit{fc}$ layer ($\in{\mathbb {R}^{400}}$) as the visual activity concept. To get the explicit clip-level activity concept, we use only the central clip activity concept.



\textbf{Query Embedding}. Same as~\cite{Gao_2017_ICCV}, we use bidirectional skip-thought~\cite{NIPS2015_5950} to encode our sentence in both normal and reverse orders. The dimension of the sentence embedding is 4,800.

\textbf{Semantic Activity Concept}. 
To get the VO embedding, we first use an off-the-shelf dependency parser~\cite{chen2014fast}. After such parsing, there are approximately 50 grammatical relations. Then we select direct object (\textit{dobj}) relational type as the predicate/object tuple. 
We only keep the two-word tuple as our VO. \ie open door, wash hands. After lemmatizing every word, we use the GloVe~\cite{pennington2014glove} word encoder which is pre-trained on the Common Crawl data to get  each word a 300-dimensional word embedding. Finally, we concatenate two 300-dimensional word embedding to get a 600-dimensional embedding as the activity concept for this sentence. Not every sentence in TACoS~\cite{regneri2013grounding} or Charades-STA~\cite{Gao_2017_ICCV} has a relational type of \textit{dobj}. Based on our results, the proportion of sentences that have \textit{dobj} for TACoS and Charades-STA are about 93\% and 68\%, respectively. If the sentence has more than one \textit{dobj}, we randomly select one. If the sentence doesn't have any \textit{dobj}, we will use a 600-dimensional zero vector instead. 

We use the scale set \{64, 128, 256, 512\} to collect our training samples with overlap 0.75 rather than 0.8 in~\cite{Gao_2017_ICCV}. So we have less training samples than~\cite{Gao_2017_ICCV}. Each training sample contains a video clip $c$ and a sentence query $q$. In each batch, the video clips and language quires can either come from the same video or different videos. The actionness score $\eta$ is used only in test. We use the pre-alignment score $\delta$ as the alignment score during training.

\begin{table}[t]
\begin{center}
\setlength\tabcolsep{4.2pt}
\begin{tabular}{l|cccc}
\hline
Method  
&\begin{tabular}{@{}c@{}}$\mathrm{R@1}$ \\ $\mathrm{IoU}$=0.7\end{tabular}
&\begin{tabular}{@{}c@{}}$\mathrm{R@1}$ \\ $\mathrm{IoU}$=0.5\end{tabular}
&\begin{tabular}{@{}c@{}}$\mathrm{R@5}$ \\ $\mathrm{IoU}$=0.7\end{tabular}
&\begin{tabular}{@{}c@{}}$\mathrm{R@5}$ \\ $\mathrm{IoU}$=0.5\end{tabular}\\

\hline\hline
Swin+Score
&8.98              &22.20              &29.38   &58.06\\
Activity only
&9.78     &22.80     &30.11  &59.29\\
w/o SAC
&9.33      &25.13      &31.02  &59.54\\
w/o VAC
&9.79      &24.03      &32.55  &60.32\\
Concat
&9.87      &24.73      &32.66  &60.80\\
ACL
&\textbf{11.23}  &\textbf{26.47}   &\textbf{33.25}  &\textbf{61.51}\\

\hline
\end{tabular}
\end{center}
\caption{Evaluation of variants of activity concepts on Charades-STA.}
\label{tbl:core_model_performance}
\end{table}

\begin{table}[t]
\begin{center}
\setlength\tabcolsep{3pt}
\begin{tabular}{l|cccc}
\hline
Method  
&\begin{tabular}{@{}c@{}}$\mathrm{R@1}$ \\ $\mathrm{IoU}$=0.7\end{tabular}
&\begin{tabular}{@{}c@{}}$\mathrm{R@1}$ \\ $\mathrm{IoU}$=0.5\end{tabular}
&\begin{tabular}{@{}c@{}}$\mathrm{R@5}$ \\ $\mathrm{IoU}$=0.7\end{tabular}
&\begin{tabular}{@{}c@{}}$\mathrm{R@5}$ \\ $\mathrm{IoU}$=0.5\end{tabular}\\

\hline\hline
Random  
&3.03   &8.51   &14.06  &37.12\\
\hline
VSA-STV~\cite{Karpathy_2015_CVPR}
&5.81   &16.91  &23.58  &53.89\\
CTRL~\cite{Gao_2017_ICCV}
&7.15   &21.42  &26.91  &59.11\\
\hline
ACL
&11.23  &26.47   &33.25  &61.51\\
ACL-K
&\textbf{12.20}  &\textbf{30.48}  &\textbf{35.13}  &\textbf{64.84}\\

\hline
\end{tabular}
\end{center}
\caption{Comparison with the state-of-the-art methods on Charades-STA.}
\label{tbl:charades_result}
\end{table}

\subsection{Evaluation Metrics}
Following~\cite{Gao_2017_ICCV}, we use ``$R@n, IoU=m$'' metric to evaluate our method in temporal activity localization. 
``$R@n, IoU=m$'' can be calculated as
$\frac{1}{N_q}\sum_{i=1}^{N_q} r(n,m,q_i)$,
where ${N_q}$ is the total number of queries, $r(n,m,q_i)$ is the alignment result for the query $q_i$ where 1 indicates correct alignment and 0 indicates wrong alignment within the $\mathit{n}$ highest scored video clips having $\mathit{IoU}$ equal or larger than $\mathit{m}$. So ``$R@n, IoU=m$'' is the averaged performance on all given queries.

\begin{table}[t]
\begin{center}
\setlength\tabcolsep{3 pt}
\begin{tabular}{l|cccccc}
\hline
Method  
&\begin{tabular}{@{}c@{}}$\mathrm{R@1}$ \\ $\mathrm{IoU}$=0.5\end{tabular}
&\begin{tabular}{@{}c@{}}$\mathrm{R@1}$ \\ $\mathrm{IoU}$=0.3\end{tabular}
&\begin{tabular}{@{}c@{}}$\mathrm{R@5}$ \\ $\mathrm{IoU}$=0.5\end{tabular}
&\begin{tabular}{@{}c@{}}$\mathrm{R@5}$ \\ $\mathrm{IoU}$=0.3\end{tabular}\\

\hline\hline

\hline
Swin+Score
&16.46      &21.70      &28.61      &39.51\\
\hline
Fusion(0.7)
&16.04      &20.94      &28.21      &38.74\\
Fusion(0.8)
&16.51      &21.82      &28.85      &39.70\\
Fusion(0.9)
&16.46      &21.67      &28.65      &39.65\\
\hline
Activity only
&14.40      &18.03      &24.03      &32.45\\
w/o SAC
&16.41      &20.79      &28.92      &38.94\\

w/o VAC
&16.63      &21.48      &28.78      &39.23\\
Concat
&16.83      &20.50      &28.97      &39.21\\ 
ACL
&\textbf{17.78}  &\textbf{22.07}  &\textbf{29.56}    &\textbf{39.73}\\

\hline
\end{tabular}
\end{center}
\caption{Evaluation of variants of activity concepts on TACoS.}
\label{tbl:tacos_ab}
\end{table}

\begin{table*}[t]
\begin{center}
\setlength\tabcolsep{10pt}
\begin{tabular}{l|cccccc}
\hline
Method  
&\begin{tabular}{@{}c@{}}$\mathrm{R@1}$ \\ $\mathrm{IoU}$=0.5\end{tabular}
&\begin{tabular}{@{}c@{}}$\mathrm{R@1}$ \\ $\mathrm{IoU}$=0.3\end{tabular}
&\begin{tabular}{@{}c@{}}$\mathrm{R@1}$ \\ $\mathrm{IoU}$=0.1\end{tabular}
&\begin{tabular}{@{}c@{}}$\mathrm{R@5}$ \\ $\mathrm{IoU}$=0.5\end{tabular}
&\begin{tabular}{@{}c@{}}$\mathrm{R@5}$ \\ $\mathrm{IoU}$=0.3\end{tabular}
&\begin{tabular}{@{}c@{}}$\mathrm{R@5}$ \\ $\mathrm{IoU}$=0.1\end{tabular}\\

\hline\hline
Random  
&0.83   &1.81   &3.28  &3.57    &7.03   &15.09\\
\hline
VSA-STV~\cite{Karpathy_2015_CVPR}
&7.56   &10.77   &15.01   &15.50   &23.92  &32.82\\
CTRL~\cite{Gao_2017_ICCV}
&13.30  &18.32  &24.32  &25.42  &36.69  &48.73\\
MCF~\cite{wu2018multi}
&12.53  &18.64  &25.84  &24.73  &37.13  &52.96\\
ACRN~\cite{liu2018attentive}
&14.62  &19.52  &24.22  &24.88  &34.97  &47.42\\

\hline
ACL
&17.78  &22.07  &28.31  &29.56  &39.73  &53.91\\
ACL-K
&\textbf{20.01}  &\textbf{24.17}   &\textbf{31.64} &\textbf{30.66}  &\textbf{42.15}  &\textbf{57.85}\\

\hline
\end{tabular}
\end{center}
\caption{Comparison with the-state-of-the-art methods on TACoS.}
\label{tbl:tacos_final}
\end{table*}

\begin{figure*}
\begin{center}
\includegraphics[width=5.5in]{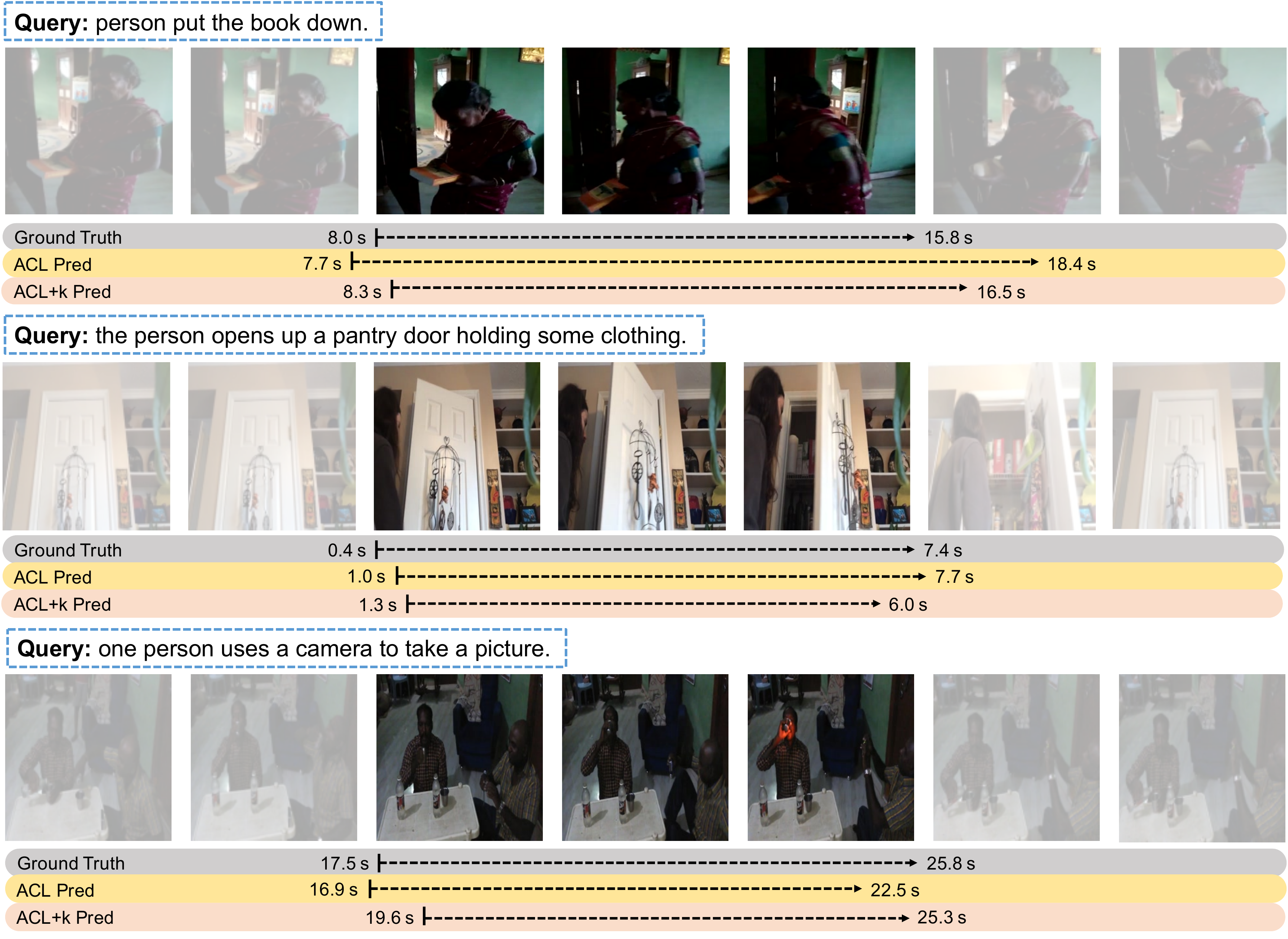}
\end{center}
\caption{Results visualization on Charades-STA. We show the ground truth, ACL results and ACL-K results in gray, yellow and pink rows, respectively. The query is on the top-left corner. Better viewed in color.}
\label{fig:charades_vis}
\end{figure*}


\begin{figure*}
\begin{center}
\includegraphics[width=5.5in]{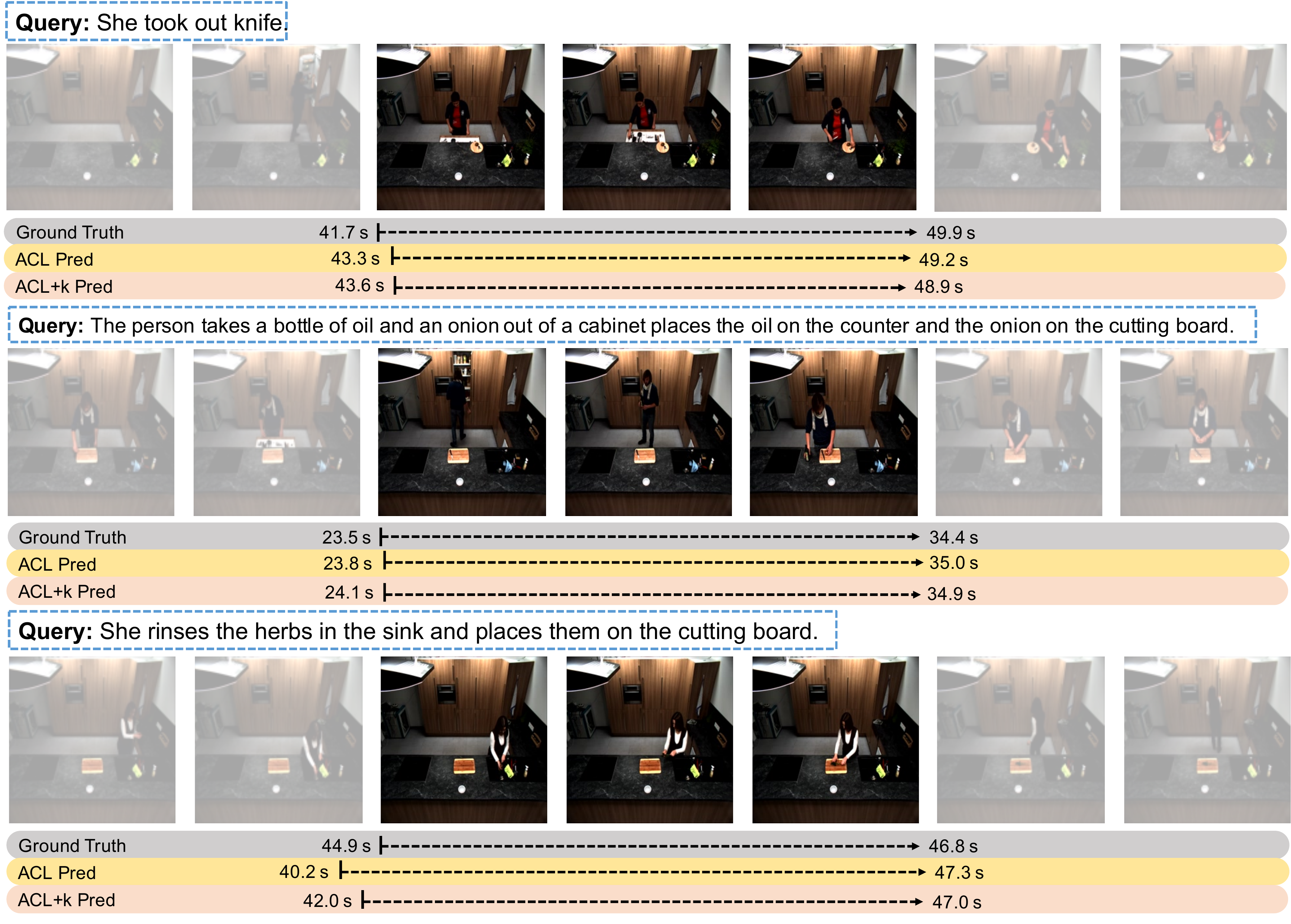}
\end{center}
\caption{Results visualization on TACoS. Better viewed in color.}
\label{fig:tacos_vis}
\end{figure*}

\subsection{Experiments on Charades-STA}
\label{subsec:charades_exp}
We evaluate our method on Charades-STA~\cite{Gao_2017_ICCV} dataset. In test, we set sliding windows length set of \{128, 256\} with $IoU=0.75$. The batch size and $\beta$ is set to 28 and 0.01, respectively. We use the Adam~\cite{kingma2014adam} optimizer with learning rate 0.005 to optimize our network.

\textbf{Actionness Score}. We compare our actionness score enhanced methods with the pure sliding window based method in~\cite{Gao_2017_ICCV}. To fairly compare our actionness score enhanced methods with CTRL~\cite{Gao_2017_ICCV}, we modify the original CTRL and made it compatible with our actionness scores.

We explore two methods to use the actionness score. The first one is to use the same sliding window as we stated above but we multiply the corresponding actionness score $\eta_i$ to each alignment score $\delta_{i,j}$ generated by CTRL (\textit{abbr.} Swin+Score). The other is to use the proposals and the corresponding actionness score $\eta_i$ (\textit{abbr.} Prop+Score). Proposals can be treated as refined sliding windows with better temporal locations for actions. From Table~\ref{tbl:score}, we can see that Prop+Score improves performance by about 2\% under ``$R@1,IoU=0.5$''. Swin+Score method also improves the localization performance. 

We compare our Swin+Score method with pure sliding window method under the metric of Average Recall-Frequency (AR-F)~\cite{TURN_2017_ICCV} in Figure~\ref{fig:ARF}. We can see that Swin+Score method has a higher average recall in all frequencies. The actionness score is able to help the model to find the more related sliding windows. In our experiments, we will use Swin+Score as default for consistency to our previous statement.

\textbf{Activity Concept}. To verify the effectiveness of activity mining, we design four system variants in this part. They are ``Activity only'', ``w/o SAC'', ``w/o VAC'', ``concat''. ``Activity only'' means we use the activity concepts only; SAC stands for Semantic Activity Concepts and VAC stands Visual Activity Concepts; ``concat'' is short for concatenation. The detailed implementations of four system variants is in Figure \ref{fig:variant}. All these variants are trained separately.


From Table \ref{tbl:core_model_performance}, we can see that ACL exceeds all four system variants by a margin. Even ``Activity only'' method exceed the previous method~\cite{Gao_2017_ICCV}. Although our ``Activity only'' method only capture the activity concepts from the sentence queries and videos, it shows us better performance than indiscriminately modling the holistic sentence and video visual information on Charades-STA~\cite{Gao_2017_ICCV}.



\textbf{Compare to State-of-the-art Methods}. For Charades-STA~\cite{Gao_2017_ICCV}, we compare our method with DVSA~\cite{Karpathy_2015_CVPR} and CTRL~\cite{Gao_2017_ICCV}. For DVSA, we use a modified version from ~\cite{Gao_2017_ICCV} to fairly compare with our methods which is short as VSA-STV in our paper. From Table~\ref{tbl:charades_result}, we can see ACL with the Sports-1M~\cite{KarpathyCVPR14} activity labels exceeds state-of-the-art-method over 3\% under the ``$R@1,IoU=0.5$'' metric. By substituting the activity label list from Sports-1M to Kinetics~\cite{kay2017kinetics}, we achieve a performance gain of over 5\%, which is short as ``ACL-K''. 




\subsection{Experiments on TACoS}
In this part, we evaluate our work on the TACoS~\cite{regneri2013grounding}. We set sliding windows length set of \{128, 256\} with $\mathrm{IoU=0.75}$ in test. All other settings are same to experiments we did in Charades-STA~\cite{Gao_2017_ICCV}. 

\textbf{Activity Concept}. We use four system variants as in Charades-STA. The performance of all variants are listed in Table~\ref{tbl:tacos_ab}. We can see that our ACL achieves the best performance to other variants. It verifies the effectiveness of our ACL on TACoS. 

Unlike the ``Activity only'' on Charades-STA~\cite{Gao_2017_ICCV}, the ``Activity only'' on TACoS~\cite{regneri2013grounding} is not better than the Swin+Score method. This can be explained by the differences between two datasets. From Table~\ref{tbl:dataset}, we can see two important differences between Charades-STA and TACoS datasets. First, the average length of queries in Charades-STA is shorter. Second, the variance of number of words in each query in Charades-STA is much smaller. These two indicate that the language queries in Charades-STA are shorter and much length-consistent than TACoS. So the ``Activity only'' method showing better performance on Charades-STA proves that our ``Activity only'' method is superior to deal with short and simple quires compared to the existing method~\cite{Gao_2017_ICCV}. 

\textbf{Video Activity Probability Late Fusion}. To compare with the proposed ACL, we conduct experiments on directly using the activity class probabilities with score fusion. 
We first compare the cosine similarity between the VO embedding of query $q_j$ with all pre-defined activity labels' embeddings. Then we retrieve the activity label $k$ that has the highest similarity score $s_{sim}$. Through the activity label $k$, we can get the probability $P^{(k)}_i$ for video clip $c_i$. $P^{(k)}_i$ can be expressed as the probability that the clip $c_i$ contains the activity label $k$. $P^{(k)}_i\times \xi_{i,j}$ is used as the fused alignment score, where $\xi_{i,j}$ is the alignment score before late fusion. To filter out the VO that is not related to any activity labels, we set threshold $\theta$ to ensure that the multiplication only happens when $s_{sim}$ is larger than $\theta$, \ie if we cannot find an activity label that is similar to VO, we don't use any video activity labels for query $q_j$. The three best results are listed as ``Fusion($\theta$)'' in Table~\ref{tbl:tacos_ab}. We build this on Swin+Score method, but the improvements to Swin+Score are small. Using activity labels instead of activity concepts does not bring improvements.



\textbf{Compare to State-of-the-art Methods}. We compare our method with previous methods in Table~\ref{tbl:tacos_final}. They are DVSA~\cite{Karpathy_2015_CVPR}, CTRL~\cite{Gao_2017_ICCV}, MCF~\cite{wu2018multi} and ACRN~\cite{liu2018attentive}. Our method shows advantage upon all of them. Our ACL with Kinetics~\cite{kay2017kinetics} activity labels (\textit{abbr.} ACL-K) exceeds the best previous method~\cite{liu2018attentive} by 5\%. Our Swin+Score method alone will improve the performance about 2\%.

\section{Conclusion}

In this paper, we addressed the problem of language-based temporal localization. We presented a novel actionness score enhanced Activity Concepts based Localizer (ACL) to localize the activities of natural language queries. We are the first to mine the activity concepts from both the videos and the sentence queries to facilitate the localization. All experimental results proved the effectiveness of our proposed method. 

\section*{Acknowledgements}
This research was supported, in part, by the Office of Naval Research under grant N00014-18-1-2050 and by an Amazon Research Award.

{\small
\bibliographystyle{ieee}
\bibliography{egbib}
}

\end{document}